\documentclass[runningheads,a4paper]{llncs}

\usepackage{amssymb}
\usepackage{graphicx}
\usepackage[usenames, dvipsnames]{color}
\usepackage{wrapfig}
\usepackage[rightcaption]{sidecap}

\usepackage{xargs}  
\usepackage[pdftex,dvipsnames]{xcolor}  
\usepackage[colorinlistoftodos,prependcaption,textsize=tiny]{todonotes}
\newcommandx{\unsure}[2][1=]{\todo[linecolor=red,backgroundcolor=red!25,bordercolor=red,#1]{#2}}
\newcommandx{\change}[2][1=]{\todo[linecolor=blue,backgroundcolor=white!25,bordercolor=blue,fancyline,#1]{#2}}
\newcommandx{\info}[2][1=]{\todo[linecolor=OliveGreen,backgroundcolor=OliveGreen!25,bordercolor=OliveGreen,#1]{#2}}
\newcommandx{\improvement}[2][1=]{\todo[linecolor=Plum,backgroundcolor=Plum!25,bordercolor=Plum,#1]{#2}}
\newcommandx{\thiswillnotshow}[2][1=]{\todo[disable,#1]{#2}}

\usepackage{url}
\urldef{\mailsa}\path|{nikhilcd,albertor}@mit.edu|    

\newcommand{\secref}[1]{Section~\ref{#1}}
\newcommand{\tabref}[1]{Table~\ref{#1}}

\newcommand{\figref}[1]{Fig.~\ref{#1}}

\usepackage{tabularx}
\usepackage{booktabs}

\newcommand{\myparagraph}[1]{\vspace{0.1in}\noindent\textbf{#1}}

\begin{document}

\mainmatter  

\title{Experimental Validation of Contact Dynamics for In-Hand Manipulation}

\titlerunning{Experimental Validation of Contact Dynamics}

%
%
\author{Roman Kolbert\inst{2} \and Nikhil Chavan-Dafle\inst{1} \and Alberto Rodriguez\inst{1}}
\authorrunning{Roman Kolbert \and Nikhil Chavan-Dafle \and Alberto Rodriguez}

\institute{Department of Mechanical Engineering,\\ Massachhusetts Institute of Technology, Cambridge, USA\\
\mailsa\\
\and Robotics and Biology Laboratory,\\ Technische Universit\"at Berlin, Berlin, Germany\\
\email{romankolbert@hotmail.com}}

%
%

\toctitle{Lecture Notes in Computer Science}
\tocauthor{Authors' Instructions}
\maketitle
\begin{abstract}
\textcolor{black}{This paper evaluates state-of-the-art contact models at predicting the motions and forces involved in simple in-hand robotic manipulations.} 
In particular it focuses on three primitive actions---\textit{linear sliding}, \textit{pivoting}, and \textit{rolling}---that involve contacts between a gripper, a rigid object, and their environment. 
The evaluation is done through thousands of controlled experiments designed to capture the motion of object and gripper, and all contact forces and torques at 250Hz.
We demonstrate that a contact modeling approach based on Coulomb's friction law and maximum energy principle is effective at reasoning about interaction to first order, but limited for making accurate predictions. We attribute the major limitations to 1) the non-uniqueness of force resolution inherent to grasps with multiple hard contacts of complex geometries, 2) unmodeled dynamics due to contact compliance, and 3) unmodeled geometries due to manufacturing defects.
\end{abstract}

\section{Introduction}
\label{sec:introduction}
Advances in computer vision and touch sensing over the last few decades have facilitated effective perception of robots and their environments. Robots can know where they are with respect to other objects and surfaces, both in contact and at a distance, which opens up opportunities to plan and control their interaction.
This paper is concerned with the experimental evaluation of state-of-
the-art contact resolution techniques in robotics that explain that interaction
in terms of predicted motions and forces. The application of choice is prehensile pushing---a form of in-hand manipulation that relies on the environment, acting as an external finger~\cite{nikhil14}, to manipulate a grasped object. 

Prehensile pushing is a complex manipulation problem where the geometries/friction/motions of the gripper, the grasped object, and the environment all play an important role in determining the resultant motion of the object.
In recent work Chavan-Dafle and Rodriguez~\cite{nikhil15b} described an algorithm to model their interaction based on classical complementarity conditions for frictional point contacts~\cite{Stewart96}, and a decomposition of complex contact geometries into rigid networks of point contacts.
Under the assumption of rigid geometries, the algorithm predicts the motion of the manipulated object along with all acting forces and torques, as it is pushed against the environment.
The main focus of this paper is to evaluate through careful experimentation the predictions by the proposed hard-contact model and by an equivalent state-of-the-art soft-contact model, in particular the physics engine MuJoCo~\cite{Todorov_mujoco}. We are specially interested in determining regions where the models produce acceptable predictions, and better understanding augmentation needs for a more realistic outcome.

The experiments, described in \secref{sec:validation}, are conducted with an accurate industrial robotic arm fitted with a parallel jaw gripper, a Vicon tracking system to capture the pose of the object, and 6 axis force-torque sensors behind all contacts to provide calibrated ground truth. We evaluate the ability of the algorithms to predict motions and force for three primitive in-hand manipulation actions: linear sliding, pivoting, and rolling. 
We observe that the models, after careful tuning, can explain to first order the overall behavior of these actions. However, limitations easily show up due to the uncontrolled variability in the execution of the actions, due to the high sensitivity of the problem to small defects in object geometries and manufacturing features, and the unmodeled or difficult-to-calibrate effects in the rigidity/compliance of contacts.

\section{Related Work}

Frictional contact has been rigorously studied for many years. Research from diverse fields of study have lead to fundamental theories and empirical models to explain the mechanics of friction \cite{Hertz1882,Bailey75,Ruina83,oden85,Han96,Urbakh2004,Autumn06}.
Starting with Hertzian contact theory~\cite{Hertz1882} for linear elastic contacts, all the way to recent nonlinear contact models of soft contacts~\cite{Urbakh2004,Ho2013}, have been used to produce simulations of the local interaction between bodies, in terms of motions, forces, and deformations.

For computation reasons, the robotics community has traditionally preferred simple Coulomb point-contact models to explain physical interaction.
A point contact model, with infinitesimal contact surface, offers only frictional forces within the tangential contact plane. 
A contact with finite geometry, modelling a softer point contact, provides also frictional torque about the contact normal. 
The relationship between the available linear and torsional friction force at a contact with finite geometry has been source of many works in robotics. Howe and Cutcosky~\cite{Howe1996} and Xydas and Kao~\cite{Xydas99} provide experimental validation of different soft contact models and approximations. These models, although proven useful for relatively hard materials, are computationally expensive; and require difficult-to-calibrate parameters or hard-to-satisfy assumptions of the distribution of pressure across the contact surface. Hence, the simple point contact model, though mostly unrealistic, is a prevalent model in robotics.

In the last few decades, the robotic manipulation community has developed a large body of work based on simple point contact models. From seminal work on dexterous manipulation \cite{Salisbury82,Fearing86} and rigid body simulations \cite{Stewart96,CherifGupta99}, to more recent work on trajectory optimization~\cite{Posa2014}, control~\cite{Tassa12}, state estimation~\cite{Yu2015}, or system identification~\cite{Fazeli2015} of systems undergoing frictional contact interactions.
Unfortunately, little attention has been given to validating the assumptions the models build on, and the realism of their predictions. In previous work~\cite{Yu2016} we studied the validity of common assumptions in planar pushing interactions.
This paper contributes with an experimental study of two models for contact resolution in the context of in-hand manipulation, and with a large dataset of carefully designed and recorded experiments.

\section{Prehensile Pushing}
\label{sec:prehensile_pushing}
\begin{figure}
\centering
\includegraphics[height=3.5cm]{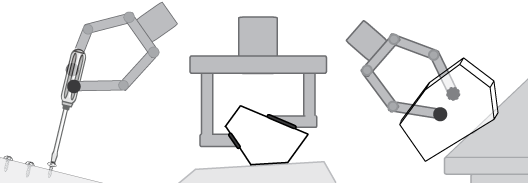}
\caption{Examples of prehensile pushes: rotating about an axis, pushing against a plane, and pivoting about an edge. (Figure from \cite{nikhil15b})}
\label{fig:examples}
\end{figure}

Our long term vision is to produce fast and reliable physical interaction between a robot and its environment. In particular we are interested in enabling prehensile pushing~\cite{nikhil15b} as a general approach to manipulation of grasped objects. Prehensile pushing addresses in-hand manipulation as a sequence of simple robust pushes that control the grasp on an object by exploiting contacts with the environment and accurate arm motions.
\figref{fig:examples} shows examples of prehensile pushes with different contact geometries and interactions.

A reliable dynamic model is a fundamental building block to plan such motions, either for search-based or optimization-based planning methods. It is crucial to that end to understand under what conditions their assumptions are reasonable, and to what degree we can trust the predictions those models generate.

In \cite{nikhil15b} we model prehensile pushing as a dynamic system composed of a rigid object in contact with the fingertips of the gripper, and the environment. In our formulation, the environment/pusher, acts as an extra finger that moves along a given trajectory and forces the object into a different grasp.
Since the motion of the environment/pusher is the reflection of the motion of a dexterous robot arm, prehensile pushing can potentially give robots unprecedented levels of dexterity.
\textcolor{black}{We model the dynamics as a set of contact forces applied on the object through possibly complex planar contacts (e.g., line, patch,...), which we model as arrays of rigidly connected point contacts.}
The resultant motion of the object, and the forces at all contacts are predicted in a time-stepping fashion as a consequence of Coulomb friction, the principle of maximum energy dissipation, non-penetration, and the motion of the pusher/environment.

In this paper, we focus on the experimental validation of three prehensile pushing primitives--linear pushing, pivoting and rolling--executed under varying experimental conditions, including griping force, pushing velocity and pushing direction.
We expect the model to be effective at reasoning about the interactions to first order.
In practice, we see that prehensile pushing with multiple contacts with complex geometries is sufficient to expose the limitations of state-of-the-art contact modeling techniques. 
These manifest in the form of non-unique valid solutions and inaccurate predictions due to unmodeled compliance and manufacturing defects in the geometry and friction of the contact surfaces.

\begin{SCfigure}
\centering
\includegraphics[height=4.5cm]{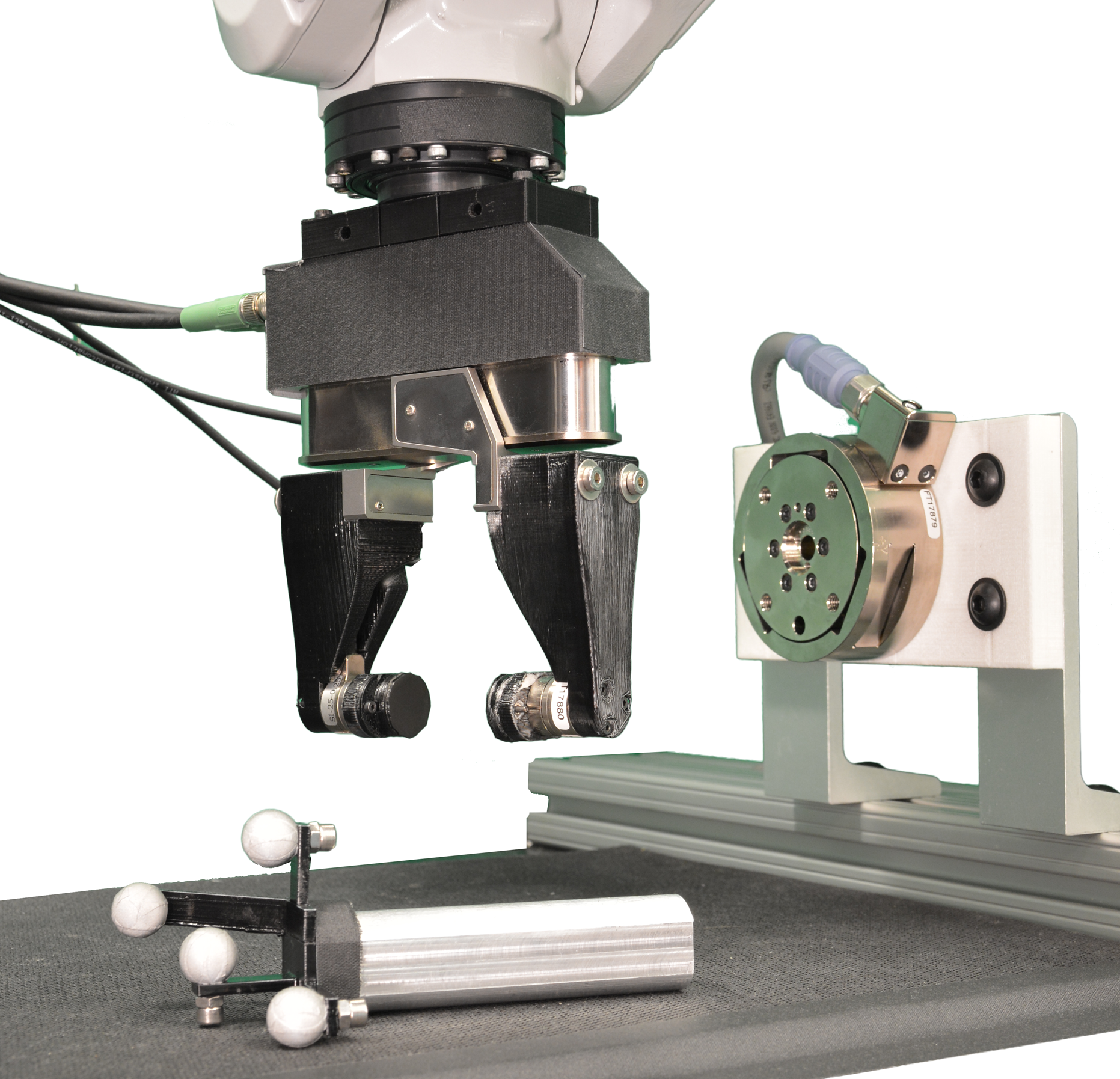}
\caption{Experimental setup with 6 axis force-torque sensors fitted to both fingertips of the gripper and in the environment. In this case, a 3D printed Vicon marker is attached to the left end of the object for accurate and fast position tracking. The setup allows us to capture ground truth for motions and contact forces at 250 Hz.}
\label{fig:setup}
\end{SCfigure}

\section{Experimental Validation}
\label{sec:validation}

\subsection{Experimental Setup}
\label{sec:setup}
\myparagraph{Setup.} \figref{fig:setup} shows the setup we use to capture the dynamics in the contact experiments proposed in \secref{sec:prehensile_pushing}. It is instrumented to capture the 6DOF pose of the object, and the 6DOF forces and torques at all contacts at 250Hz. We use an accurate ABB IRB120 robotic arm and a force-controlled parallel-jaw gripper to grasp the objects and push them against the environment. We use a Vicon motion tracking system with Bonita cameras to capture the pose of the object. The fingertips of the gripper are instrumented with ATI Nano17 F/T sensors and the environment with an ATI Gamma F/T sensor. For improved accuracy, we run a basic calibration routine for all F/T sensors, to eliminate the intrinsic offsets if any.

\myparagraph{Objects.} For the experiments in this work, we use the five objects in \tabref{tab:objects}, adding variability in materials, geometry, and weight.

\begin{table}
  \caption{Objects used in experiments. Physical properties.}
  \label{tab:objects}
	\centering
	\begin{tabular}{|r|l|l|r|r|r|}
         \hline
          & \textbf{Shape} & \textbf{Material} & \textbf{Length} (mm) & \textbf{Side} (mm) & \textbf{Mass} (g)\\ \hline
          \texttt{\ obj1\ } & cylinder  & aluminum 6061 & 100 & 25 & 158 \\ \hline  
          \texttt{\ obj2\ } & cylinder  & ABS & 100 & 25 & 72.5 \\ \hline
          \texttt{\ obj3\ } & cylinder with flat faces  & aluminum 6061 & 100 & 25 & 145 \\ \hline
          \texttt{\ obj4\ } & square prism  & aluminum 6061 & 100 & 25 & 200.5 \\ \hline
          \texttt{\ obj5\ } & square prism  & ABS & 100 & 25 & 93.8 \\ \hline
    \end{tabular}
\end{table}

\myparagraph{Contacts.} We designed exchangeable fingertips of different contact geometry (point, line, and circular) to be attached to the F/T sensors in the fingers and environment. These are 3D printed in a hard plastic material and covered with a thin layer of hard rubber which provided a good compromise between hardness, high friction and abrasion resistance. 
\textcolor{black}{We identify the coefficient of friction at contacts with a small amount of experimental data, such that simulations and experiments yield similar results.} The coefficients for the different contacts used in the experiments are in the range $[0.35,0.5]$ at fingers and $[0.2,0.3]$ at the pusher/environment. The rest of the data is used for validation.

\subsection{Experimental Results}
\label{sec:results}
\subsubsection{Linear Pushing.}
\label{sec:linpush}
\begin{figure}
\centering
\includegraphics[height=2.5cm]{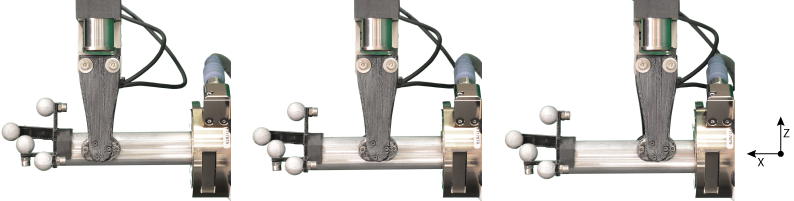}\\
\vspace{0.2cm}
\includegraphics[width=11cm]{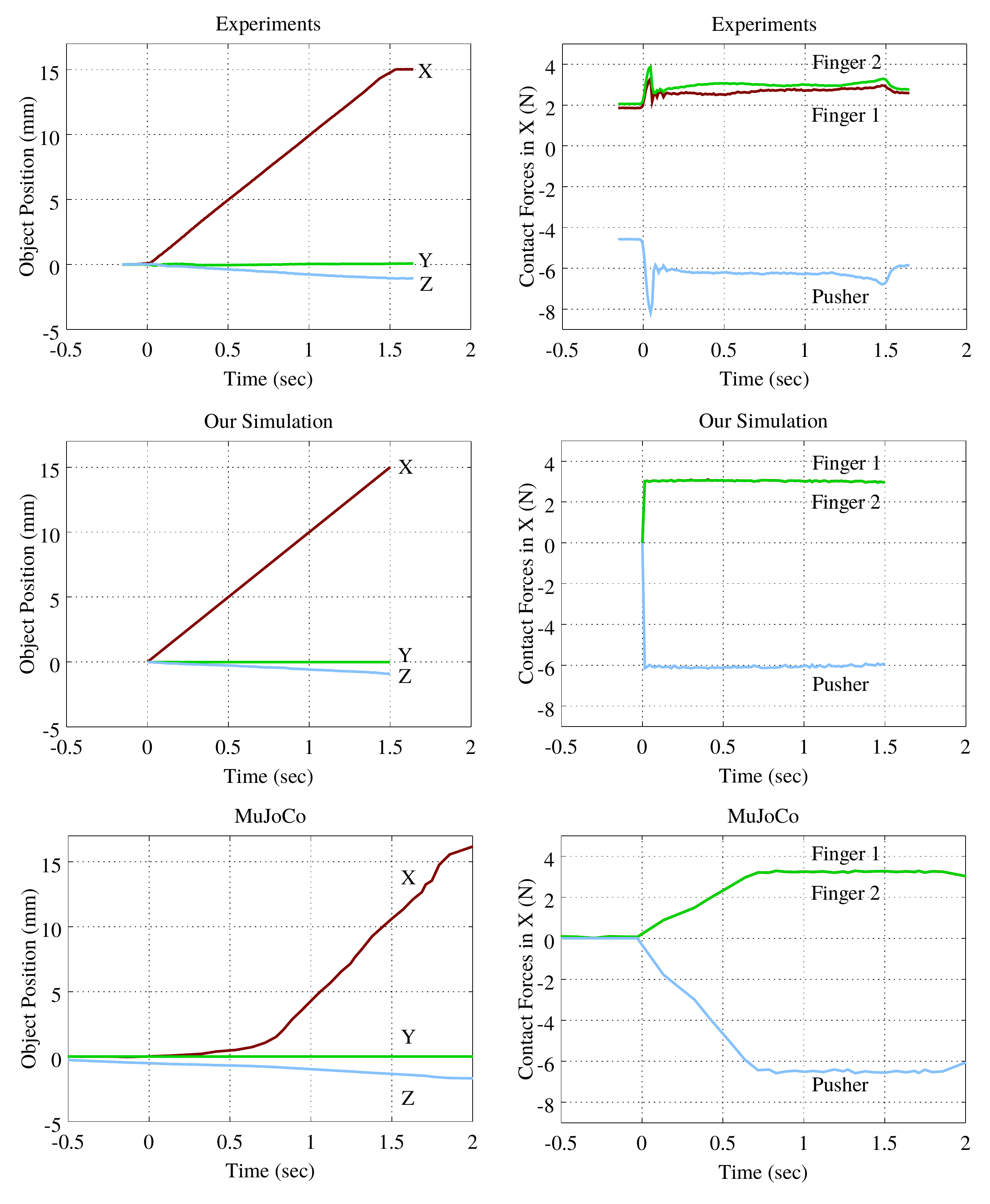}
\caption{Example of a linear push with gripping force of $20$ N, pushing velocity of $10$ mm/sec and $0$ degree slope to the horizontal. Note the motion of the object in the -Z direction, as the gripper moves straight in -X direction. The plots show the  experimental data captured for motions on the left and force on the right (top), and simulated data with the hard-contact model in~\cite{nikhil15b} (mid) and with the soft-contact model in  MuJoCo~\cite{Todorov_mujoco} (bottom). Nominally, the pushing action starts at $0$~sec and ends at $1.5$~sec.}
\label{fig:linplots}
\end{figure}

In this experiment, a prismatic shaped object is pushed against a flat face of the force-torque sensor along a straight line (top of \figref{fig:linplots}). We chose the fingertip contacts to be circular flat contacts of diameter $20$ mm. We run multiple straight pushes by changing the gripping force $\in\ \{20,22,25,27,30,\\
32,35\}$ N, the pushing velocity  $\in\ \{10,15,20,25\}$ mm/sec and the slope of the straight push with respect to the horizontal $\in\ \{-20,-10,0,10,20\}$. We performed three runs for each combination collectively yielding $420$ each for objects \texttt{obj3}, \texttt{obj4}, and \texttt{obj5}. The complete raw data and helper files to parse it are available in the online repository~\cite{datalink}

Figure~\ref{fig:linplots} shows results for one of those experiments and compares with the predictions by the two models of choice. The hard-contact model correctly predicts the motion of the object along the pushing direction (X) with no motion perpendicular to the flat faces of the fingertips (Y). It also predicts the downward sliding motion (Z) of about $1$~mm, due to gravity pulling on the object.
The experimental data shows force peaks at the fingertips and pusher right at the beginning of the motion. These peaks can be attributed to an impact phase, and to the known differences between kinetic and static friction, which are not considered in any of the two models. Note also that the predicted pushing force of $6$~N is very close to the force observed in experiments, $6.2$~N.

In the series of experiments we also observe that increasing the gripping
force, leading to a higher pushing force required to move the object and to a
diminished falling motion along axis Z. Above $25$~N of gripping force, the object barely slides down, a behavior correctly predicted by the hard contact model. Changes in the pushing velocity showed no significant change either in the forces at contacts or in the object motion.

\subsubsection{Pivoting.}
\label{sec:pivot}

\begin{figure}
\centering
\includegraphics[height=3.25cm]{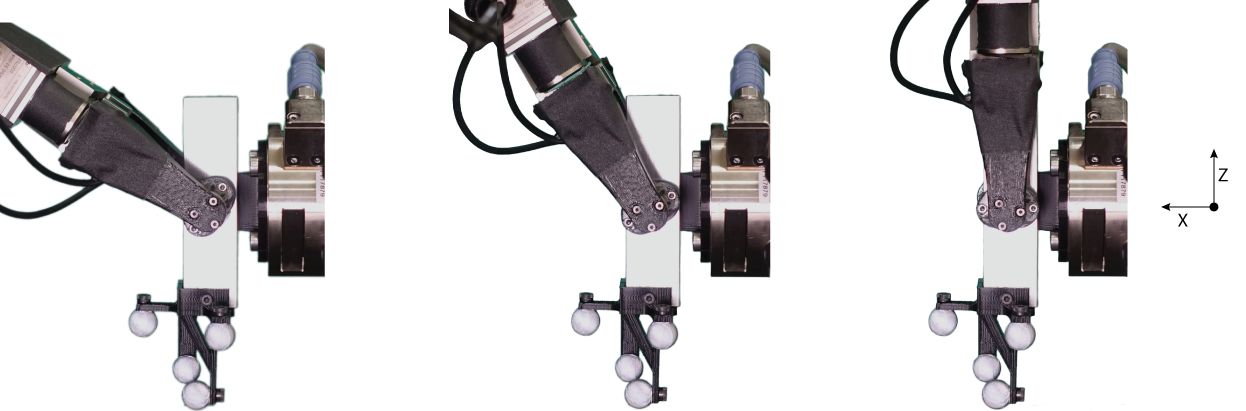}\\
\vspace{0.2cm}
\includegraphics[width=11cm]{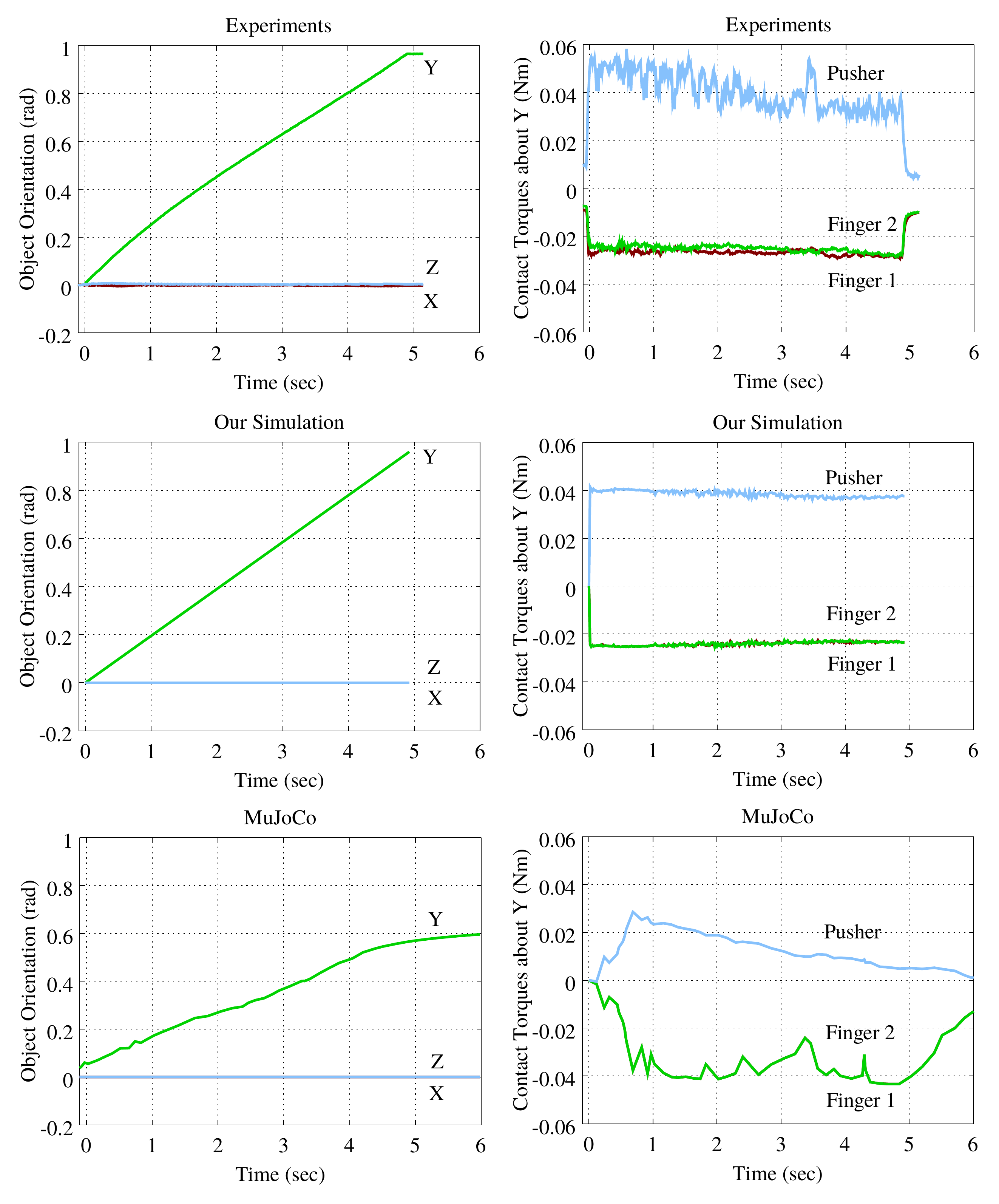}
\caption{Example of pivoting  with grasping force of $20$ N, pushing velocity of $10$ deg/sec and no pusher offset. Experimental data (top) Simulation data (mid) and MuJoCo data (bottom) for object orientation and contact torques at the contacts about the axis of pivoting for one of the pivoting experiments}
\label{fig:pivotplots}
\end{figure}

In this experiment the griper holds a prismatic object and pivots it against a line contact. Contacts at fingertips are again flat contacts of diameter $20$~mm and that with the external feature is a line contact of length $28$~mm. We conducted multiple experiments by changing the grasping force $\in\ \{20,22,25,27,30,32,35\}$~N, the angular velocity of pivoting about fingers $\in\ \{2.5,5,10,15,20\}^{\circ}$/sec and offset of the pusher contact location from the center of the object $\in\ \{0,5,10,15,20,25\}$ mm. We performed three runs for each combination of the parameters collectively yielding $630$ each for objects \texttt{obj4} and \texttt{obj5}.

Figure~\ref{fig:pivotplots} shows an example of a pivoting experiment, as well as the results of object orientation and torques as observed experimentally and predicted by the two models of choice. Experiments and the two models agree that the object will remain virtually stationary while pivoting against the external line. 
The hard contact model also gives a good prediction of the torque experienced at the fingertips and pusher about the axis of pivoting. 

In general, we find agreement to first order between the analytical predictions from the hard model and experimental observations.
The pushing force and torque experienced by the pusher increase as the gripping force increase or/and when the velocity of the push increases. The hard model, and to some degree the soft contact model, capture these trends. For the hard contact model, the predicted torques differ from the experimental values by as much as $0.005$~Nm at the fingers and $0.01$~Nm at the pusher. Note that the sensor uncertainty is $0.003$~Nm at the fingers and $0.037$~Nm at the pusher.
We also observe a consistent slight decrease in the torque at the pusher as the object rotates, phenomenon which is not predicted by the hard-contact model, so could be due to unmodeled contact compliance.
%


\subsubsection{Rolling. }
\label{sec:roll}

In this experiment the gripper forces a cylindrical object to roll on a flat platform coated with silicone rubber of high coefficient of friction. In our setup, the grasping force required to successfully roll the object while retaining the grasp on the object is predicted by the hard-contact model to be below $5$~N. It was challenging to reliably perform these experiments with our gripper which is designed to be operated above 5N force.
In this experiment, we do not have a force-torque sensor attached to the external contact. We conduct multiple experiments by changing the grasping force $\in\ \{3,4,5,6,7,8,9,10\}$~N and robot velocity $\in\ \{5,10,15,20,25\}$~mm/sec, and performed three runs of each configuration, for a total of $240$ experiments for objects \texttt{obj1} and \texttt{obj2}.

\begin{figure}[h]
\centering
\includegraphics[height=3.25cm]{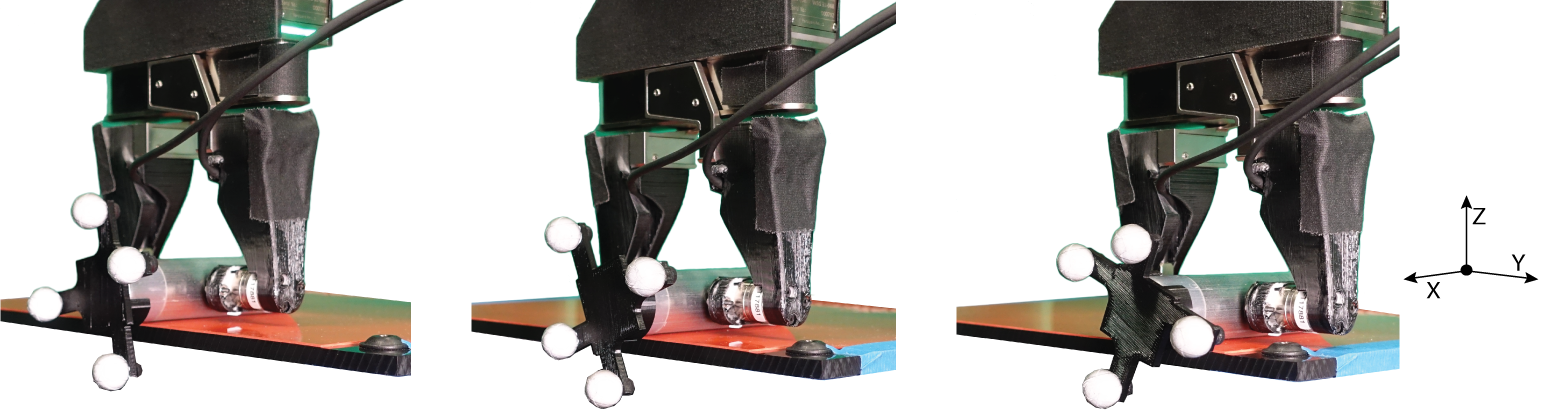}\\
\vspace{0.2cm}
\includegraphics[width=9.25cm,angle=-90]{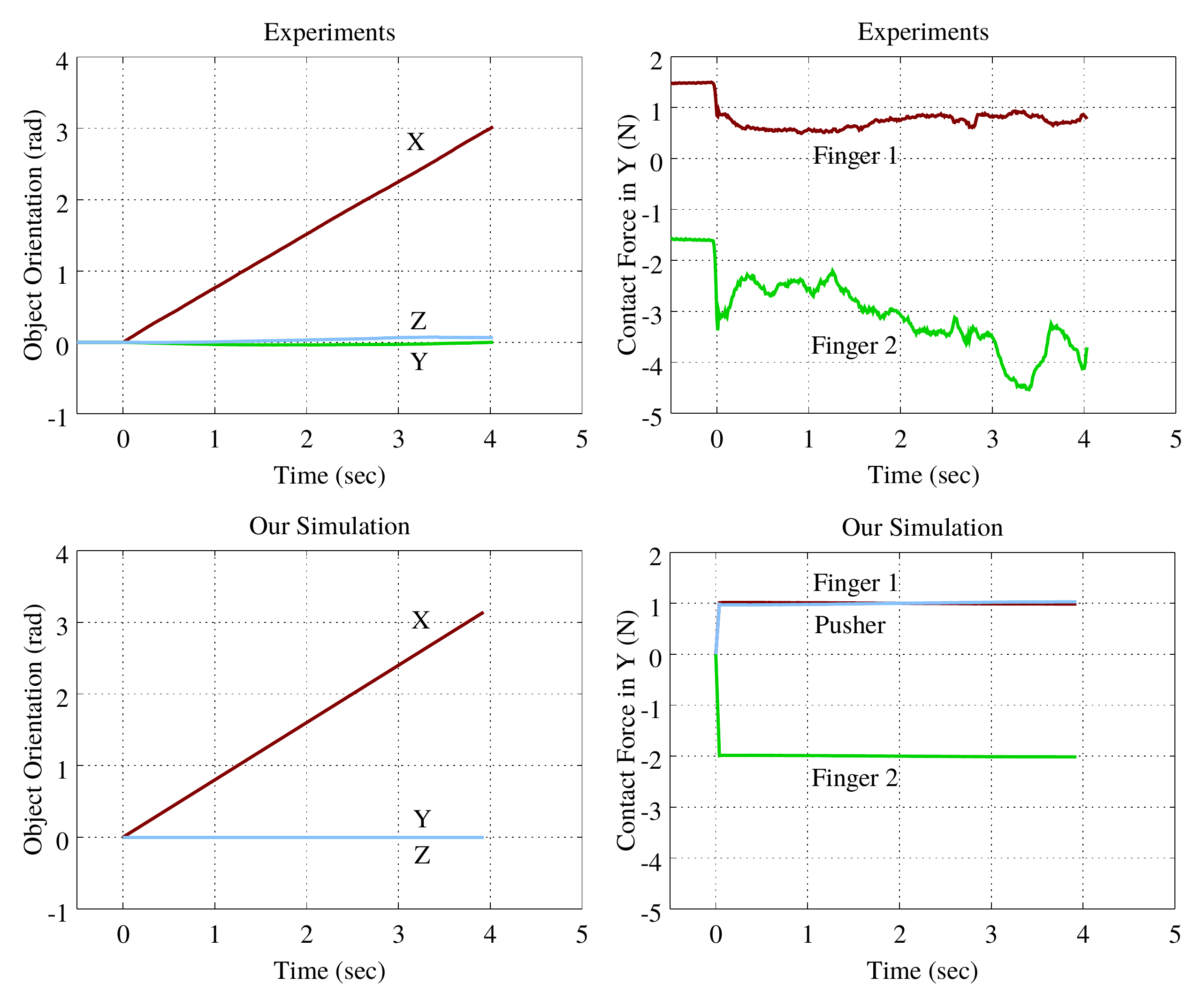}
\caption{Example of rolling with grasping force of $3$N and pushing velocity of $10$mm/sec. Three instants from left to right show the beginning of the push, intermediate position and end of the push.}
\label{fig:rollplots}
\end{figure}

Figure~\ref{fig:rollplots} shows the object orientation as seen from the gripper and the forces at the fingers along X and Z directions, when the object is rolled with a gripping force of $3$~N.
The hard-contact model predicts that, while the object is pushed in Y direction, the forces experienced at the fingers are balanced - while the finger pushing the object experiences an extra force, equal to half of the pushing force exerted by the platform, the finger in front experiences less force by the same amount. 
We see a similar behaviour in experimental data; but with a worse fitting than that with the previous experiments. This is likely due to poor force control of the gripper, given it is near its operating limit yielding momentary slack and tightening of the grasp as the object rolls.

An important caveat here is that the simulations are consistent with experimental data only if the coefficient of friction of the silicone surface is assumed to be very high (close to $2.5$). The unavailability of ground truth forces at the external contact made further analysis difficult. A further investigation of this behaviour would be required to discern if it is due to sticking effect of silicone, unstable gripper control, or other unmodelled effects, such as compliance.

\section{Observations and Discussion}
The experiments show that the studied models provide useful predictions to first degree. If not to make very accurate predictions, at least they provide accurate trends.
We conclude by discussing important misgivings of the models, and directions for future work.

\begin{figure}[t]
\centering
\includegraphics[width=9.9cm, angle=-90]{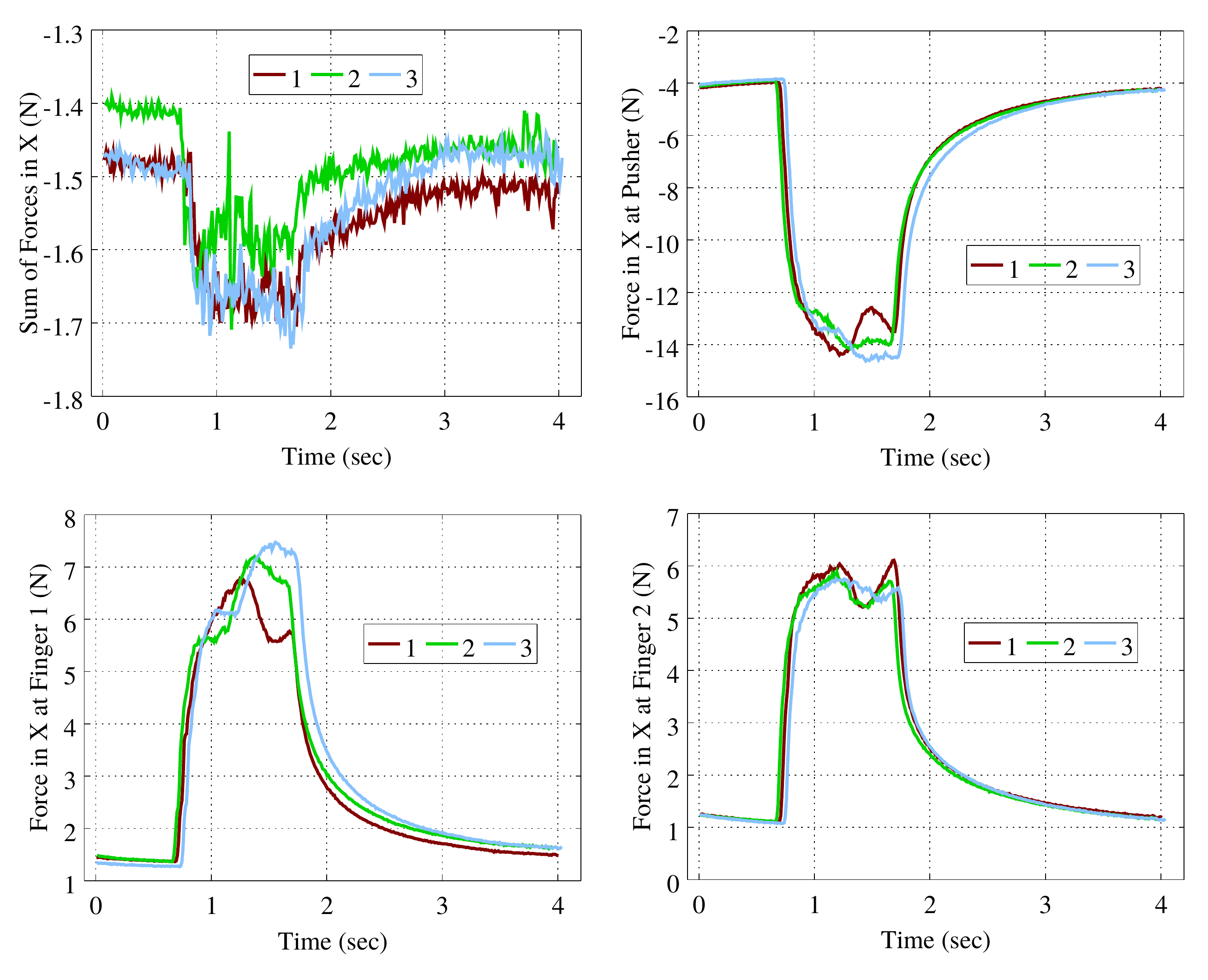}
\caption{\textcolor{black}{Three runs of linear pushing with same parameters and initial conditions. Sum of forces along x (top left), forces along x at pusher contact (top right), forces along x at finger 1 (bottom left), forces along x at finger 2 (bottom right)}}
\label{fig:var_plots}
\end{figure}

\subsection{Variability in the Experimental Data}
Figure~\ref{fig:var_plots} shows three runs of an identical linear pushing experiment. They are similar, but clearly not identical. This variability is important, and should be further quantified.
For reference, the plot on the top left shows the sum of forces in the X direction for each of those three runs. \textcolor{black}{Newton's second law says that the sum should be close to zero when the object is not moving or if the push is slow.} Note that even before the push, the forces sum up to approximately $-1.5$ N for all three runs. The two sensors in the fingers have an uncertainty of $0.25$ N and the sensor in the environment an uncertainty of $0.75$ N along the X axis. These values, together with possible small miss-alignments at  contacts could justify the experimental error.

Within the first second, the plot shows a static noise of the sensors of approximately $0.02$~N. The noise increases then between time 1 sec and 2 sec, which is the period of the push. A transition between sticking and sliding takes place. This effect was experienced specially for small velocities ($5$ mm/s) and high gripping forces($30$ N). 
 
\subsection{Multiple Valid Solutions}
\label{sec:multisolutions}

The sensor data shows a variability that cannot be just attributed to sensor noise ($0.02$N) or uncertainty ($0.25$ N). In Fig.\ref{fig:var_plots} the forces differ at some points by almost $2$N. These cases are very well correlated with the experiments where we observed discrepancies between the local contact forces and torques predicted by our simulator and those observed experimentally.
\textcolor{black}{However, as discussed in the previous section, the net wrench generated on the object as a result of these local forces and torques was found to be sufficiently close for the three runs, pointing to the possibility of high sensitivity of local forces.}
In general, we found that, there is no simple deterministic mapping between motions and forces, with forces varying by as much as $20$\%. This effect is specially related to hard surface contacts and the sensitivity to the pressure distribution, and it limits the ability of models to make accurate predictions of motions and forces. 

\subsection{Hard Vs. Soft Contacts}

\textcolor{black}{In this paper we compared our proposed hard-contact model with MuJoCo~\cite{Todorov_mujoco}, a fast physics engine designed for model-based optimization. The plots for Mujoco in Fig.\ref{fig:linplots} show that the object starts sliding down even before making contact with the external pusher.} In fact, we found it was very difficult to set up the simulation environment so that did not happen. \textcolor{black}{The soft constraints in MuJoCo make the solver very fast and useful for many applications, but also limit the prediction accuracy, especially when rigid contacts and transitions in sticking and slipping at contacts are involved.} To make the pusher motion control stable, we had to introduce damping which leads to the slower increase of contact forces.

\textcolor{black}{Equally, as seen in Fig.\ref{fig:pivotplots} for a pivoting experiment, the object motion predictions proved less realistic in MuJoCo (rotation of about 0.6 vs 1 radian)}. The soft contact model does not force the object to maintain line contact, which perturbs the natural stability of the pivoting action against a hard contact. The torques at the pusher contact differ by $0.02$N to $0.03$N from the experimental data. In contrast with the hard-model, it predicts a decreasing torque at pusher contact, which the experimental data also shows. We hypothesize this is due to actual compliance in the real contact.
The effects of the soft model can be reduced by increasing the stiffness that MuJoCo uses to impose the contact constraints. However, above a certain threshold, the system becomes unstable and noisy. Spending more time with the simulator, and better understanding how to tune it, could potentially yield better results.

\end{document}